\documentclass[11pt]{article}

\usepackage[preprint]{acl}

\usepackage{times}
\usepackage{latexsym}
\usepackage{amsmath}
\usepackage{booktabs}
\usepackage{subcaption}
\usepackage{graphicx}
\usepackage{tcolorbox}
\usepackage{xcolor}
\usepackage{algorithm}
\usepackage{algpseudocode}
\usepackage{float}
\usepackage[T1]{fontenc}
\usepackage[utf8]{inputenc}
\DeclareUnicodeCharacter{2194}{\ensuremath{\leftrightarrow}}
\DeclareUnicodeCharacter{2192}{\ensuremath{\rightarrow}}
\DeclareUnicodeCharacter{2013}{--}
\DeclareUnicodeCharacter{201C}{``}
\DeclareUnicodeCharacter{201D}{''}

\usepackage[utf8]{inputenc}

\usepackage{microtype}

\usepackage{inconsolata}

\usepackage{graphicx}

\title{
    \begin{minipage}[c]{0.10\textwidth} 
        \includegraphics[width=\linewidth]{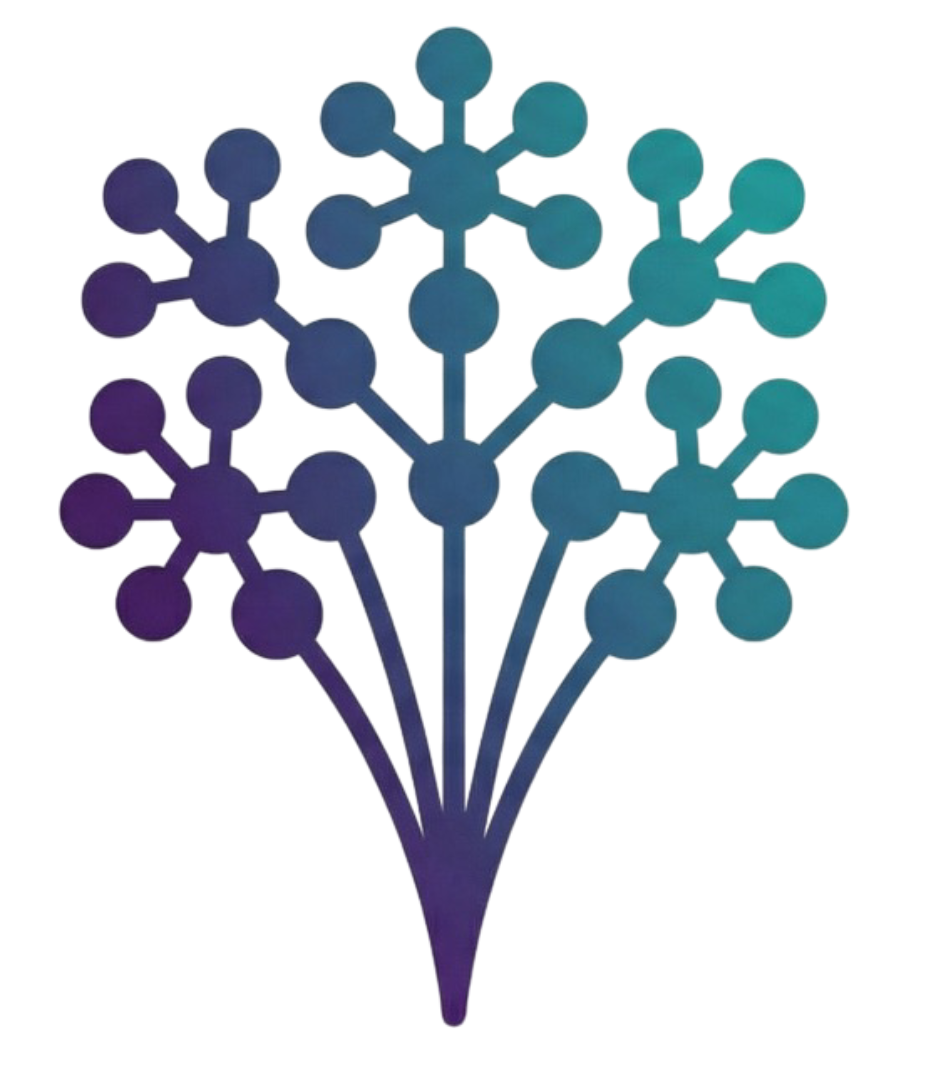}
    \end{minipage}%
    \hfill 
    \begin{minipage}[c]{0.85\textwidth}
        \raggedright 
        \bfseries    
        \Large       
        Beyond Cosine Similarity: Taming Semantic Drift and Antonym Intrusion in a 15-Million Node Turkish Synonym Graph
    \end{minipage}
    
    \vspace{-0.1cm} 
}

\author{
  Ebubekir Tosun \\
  \texttt{ebubekirsoftware@gmail.com} \\
  \And
  Mehmet Emin Buldur \\
  \texttt{mehmeteminbuldur@outlook.com} \\
  \AND  
  Özay Ezerceli \\
  \texttt{ezerceli804@gmail.com} \\
  \And
  Mahmoud ElHussieni \\
  \texttt{mahmoud.elhussieni@outlook.com} 
}

\begin{document}
\maketitle
\begin{abstract}
Neural embeddings have a notorious blind spot: they can't reliably tell synonyms apart from antonyms. Consequently, increasing similarity thresholds often fails to prevent opposites from being grouped together. We've built a large-scale semantic clustering system specifically designed to tackle this problem head-on. Our pipeline chews through 15 million lexical items, evaluates a massive 520 million potential relationships, and ultimately generates 2.9 million high-precision semantic clusters. The system makes three primary contributions. First, we introduce a labeled dataset of 843{,}000 concept pairs spanning synonymy, antonymy, and co-hyponymy, constructed via Gemini~2.5--Flash LLM augmentation and verified using human-curated dictionary resources. Second, we propose a specialized three-way semantic relation discriminator that achieves 90\% macro-F1, enabling robust disambiguation beyond raw embedding similarity. Third, we introduce a novel soft-to-hard clustering algorithm that mitigates semantic drift preventing erroneous transitive chains (e.g., \emph{hot} $\rightarrow$ \emph{spicy} $\rightarrow$ \emph{pain} $\rightarrow$ \emph{depression}) while simultaneously resolving polysemy. Our approach employs a topology-aware two-stage expansion--pruning procedure with topological voting, ensuring that each term is assigned to exactly one semantically coherent cluster. The resulting resource enables high-precision semantic search and retrieval-augmented generation, particularly for morphologically rich and low-resource languages where existing synonym databases remain sparse.
\end{abstract}

\section{Introduction}
The foundational assumption underlying most neural semantic search systems that high cosine similarity between embedding vectors correlates with synonymy is demonstrably false. Embedding models trained on contrastive objectives \cite{Wang_Isola_2020} or word2vec-style objectives \cite{Levy_Goldberg_2014} optimize for capturing distributional similarity, which conflates multiple semantic relationships into a single continuous measure \cite{Elekes_Schaeler_Boehm_2017,Schneidermann_Hvingelby_Pedersen_2020}. Antonyms and synonyms alike often occupy proximate regions in embedding space \cite{Scheible_Walde_Springorum_2013,Rizkallah_Atiya_Shaheen_2020,Samenko_Tikhonov_Yamshchikov_2020} because they appear in nearly identical linguistic contexts \cite{Liu_Liu_Chen_2017, Chaurasia_McDonnell_2016}, merely differing in implicit negation or opposition. This issue is especially pronounced when synonym databases are built by applying a high similarity cutoff (e.g., cosine similarity $\geq 0.85$), which often admits false positives such as antonyms or broadly related terms.

At scale, two factors further aggravate semantic ambiguity in graph clustering. First, modularity-based methods such as Louvain \cite{blondel2008louvain} and Leiden \cite{traag2019leiden} impose hard partitions, forcing polysemous words into a single cluster. For example, the Turkish word yüz (“face” / “100”) cannot be represented in multiple senses simultaneously, which either removes valid synonym links or collapses unrelated regions into noisy clusters. Second, semantic drift arises from transitivity: chains of locally plausible links can connect semantically distant concepts (e.g., Sıcak “hot” → Acı “spicy” → Üzüntü “pain” → Depresyon “depression”), and standard clustering methods (e.g., connected components or hierarchical agglomeration) typically cannot distinguish such spurious paths from genuine synonym structure.

These difficulties are amplified in morphologically rich, non-English settings. Turkish, in particular, lacks large machine-readable synonym resources comparable to WordNet \cite{miller1995wordnet} or BabelNet \cite{navigli2012babelnet}. Existing resources suffer from two fundamental limitations: (1) they are \textit{sparse}, containing insufficient coverage of domain-specific terminology (particularly in legal, medical, and technical domains); (2) they are \textit{brittle}, often created through heuristic rule-based systems or simple distributional similarity thresholds without explicit disambiguation of semantic relationships. The construction of a Turkish synonym resource at scale requires overcoming both antonym intrusion (false positives) and semantic drift (transitivity artifacts).

This paper contributes three substantial advances addressing these challenges:

\paragraph{A Large-Scale Disambiguated Training Dataset}
We present a rigorous methodology for generating 843,000 labeled semantic relationship pairs (Synonym, Antonym, Co-hyponym classes) through LLM-augmented synthesis paired with high-confidence dictionary extraction. We generated this dataset using Gemini 2.5-Flash \cite{comanici2025gemini25pushingfrontier} for about \$65 which is pretty cost-effective, all things considered. It serves a dual purpose: it's a novel research contribution in its own right, and it provides the training foundation for our discriminator model. The dataset strikes a balance between synthetic LLM-generated pairs (827,000 instances) and carefully curated dictionary entries (16,000 instances). What's especially useful here is that we've established a reproducible pipeline that anyone can use to generate supervised semantic relationship data for any language as long as you've got sufficient terminological resources to work with.

\paragraph{A Specialized Multi-Class Discriminator}
Instead of just tweaking unsupervised similarity thresholds and hoping for the best, we train explicit three-way classifiers that can confidently tell synonyms apart from antonyms and co-hyponyms. Our best model hits 90\% F1-macro with confidence scores above 0.98 on clear-cut cases, and we've validated that it shows solid cross-model consensus in practice. Think of this discriminator as a strict semantic gatekeeper which it only lets through high-confidence synonym predictions (confidence $\geq 0.70$) while actively filtering out antonyms and those murky low-confidence co-hyponyms.

\paragraph{A Novel Disambiguation-Aware Clustering Algorithm} We propose a two-stage graph clustering procedure (Expansion followed by Reduction/Pruning) that departs from standard community detection.

In the Expansion stage, we apply soft clustering via overlap-based neighbor inclusion, allowing terms to participate in multiple candidate clusters to reflect polysemy. In the Reduction stage, we resolve ambiguity through topological voting, a three-level hierarchical scheme (majority rule, specificity preference, and a deterministic tiebreak) that assigns each term to a single best cluster while preserving recall. The method also mitigates semantic drift by seeding clusters with high-confidence pairs, sorting candidates by confidence, and enforcing intersection-ratio thresholds ($\geq 0.51$) to prevent weak transitive chains from inflating clusters.

At scale, the pipeline processes $\sim15$ million unique terms, evaluates 520 million candidate synonym links through successive filters, and yields 2,905,071 final clusters (median size 3, mean 4.58, maximum 86). The resulting synonym graph enforces symmetry and conflict resolution, and it favors standardized domain terminology over noise such as OCR artifacts or informal abbreviations during parent selection.

\section{Related Work}

Distinguishing true synonymy from broader semantic relatedness remains a well-known limitation of distributional semantics \cite{Mohammad_Hirst_2012, McCarthy_Keller_Navigli_2010, Walde_2013}. Building on the distributional hypothesis \cite{harris1954distributional,Faruqui_Dyer_2014}, vector-space models infer meaning from contextual co-occurrence. Although this reliably captures general semantic proximity \cite{Baroni_Lenci_2010,Goyal_Jauhar_Li_Sachan_Srivastava_Hovy_2013}, it also creates systematic ambiguity: words that share contexts are embedded close together even when they stand in relations such as equivalence, opposition, or categorical relatedness \cite{Bertolotti_Cazzola_2024}. Consequently, antonyms (e.g., hot vs. cold) and co-hyponyms (e.g., cat vs. dog) can appear as similar as genuine synonyms under cosine similarity \cite{Plas_Tiedemann_2006,Yih_Zweig_Platt_2012}.

\citet{mohammad2013computing} quantified this effect, showing that standard similarity measures fail to represent relational polarity and can exhibit simultaneous similarity and contrast. Later studies report that the same limitation persists in transformer embeddings, even when trained with large-scale contrastive or retrieval objectives \cite{Etcheverry_Wonsever_2023}. Thus, despite architectural progress, these models primarily encode distributional overlap rather than semantic equivalence, which makes synonym extraction via simple similarity thresholding unreliable \cite{Garcia_2021}.

\subsection{Constraint-Based Approaches to Antonymy}

One influential line of research sought to address antonym intrusion by modifying embedding spaces using explicit lexical constraints. Counter-fitting \cite{mrksic2016counter} introduced a post-training adjustment procedure that pulls known synonyms closer together while pushing antonyms apart. Variants of this approach have shown strong improvements in intrinsic evaluation tasks and downstream lexical inference. However, their effectiveness is tightly coupled to the availability of large, high-quality lexical resources \cite{Dunietz_Levin_Carbonell_2013} such as WordNet \cite{miller1995wordnet} or PPDB \cite{ganitkevitch2013ppdb}.

For morphologically rich and comparatively low-resource languages, this dependency presents a serious bottleneck. Turkish, in particular, exhibits extensive inflectional and derivational productivity, yielding a combinatorial explosion of valid word forms that are sparsely represented in manually curated lexicons \cite{Kırkıcı_Clahsen_2013,Durrant_2013}. Existing resources such as Turkish-English Parallel TreeBank \cite{ehsani2018kenet} and KeNet \cite{bakay2021trwordnet} are largely constructed through translation-based projection or limited expert annotation, resulting in insufficient coverage for large-scale semantic modeling. As a consequence, constraint-based refinement methods do not scale effectively to the lexical breadth required for comprehensive Turkish synonym graphs.

\subsection{Graph-Based Clustering and Semantic Drift}

Many synonymy methods treat the task as graph clustering, with words as nodes and semantic proximity as edges, but they are vulnerable to semantic drift: transitive chaining of valid local links can yield globally incoherent synonym sets. Standard community detection (e.g., Louvain; \cite{blondel2008louvain}, Leiden; \cite{traag2019leiden}) further aggravates this by enforcing hard partitions that ignore polysemy, often merging unrelated senses or discarding legitimate links. Sense-aware approaches such as Watset \cite{ustalov2017watset} mitigate this by inducing senses before clustering, but the repeated local clustering steps become computationally impractical at very large scale.

\subsection{Positioning of the Present Work}

We address these issues with an integrated approach that combines semantic disambiguation, large-scale supervision, and topology-aware clustering. Instead of globally reshaping embeddings or depending on fixed lexical constraints, we train an explicit semantic relation discriminator on a large disambiguated dataset, enabling direct filtering of synonym candidates rather than equating high similarity with synonymy.

At the graph level, we avoid both strict hard partitioning and full sense induction. Our soft-to-hard clustering strategy (Appendix~\ref{app:algorithm}) allows temporary overlap to reflect polysemy, then applies a deterministic topological voting step to resolve ambiguity. This retains key advantages of sense-aware clustering while remaining scalable.

Unlike prior work, we target morphologically rich, low-resource settings. Using LLM-augmented supervision and confidence-based graph construction, we show that high-precision synonym graphs can be built without exhaustive manual lexicons, even at 15 million nodes.

\section{Methodology}
Our system is organized as a sequence of seven processing stages, each designed to address a concrete challenge encountered when building a large-scale synonym graph that is robust to antonym intrusion. Rather than treating the pipeline as a monolithic procedure, we describe each stage separately to clarify its role and design rationale. In particular, Phases~3--5 introduce the core methodological contributions of this work, where semantic disambiguation and drift-aware clustering are explicitly enforced.

\subsection{Phase 1: Embedding Model Training and Initial Data Generation}

We adopt a dual-track approach for initial data generation, combining pre-trained multilingual embeddings with LLM-augmented synonym pair synthesis.

\paragraph{Embedding Model Specialization}
Our primary semantic encoder is initialized from multilingual-e5-large \cite{wang2024multilingual}, an XLM-RoBERTa architecture trained on large-scale multilingual entailment and retrieval datasets. We fine-tune this model using contrastive learning on Turkish synonym pairs, employing CachedMultipleNegativesRankingLoss with temperature parameter $\tau = 0.07$. The loss function is formalized as:

\begin{equation}
\mathcal{L}_{\text{contrastive}} = -\log \frac{\exp(\text{sim}(q, p^+)/\tau)}{\sum_i \exp(\text{sim}(q, p_i)/\tau)}
\end{equation}

where $\text{sim}$ denotes cosine similarity between query embeddings and positive/negative examples, with cached negatives drawn from earlier batch iterations to improve sample efficiency. Model hyperparameters include maximum sequence length of 512 tokens, hidden dimension 1024, and L2 normalization of output embeddings. This training produces embeddings specifically optimized for synonym pairs rather than generic semantic similarity.

\paragraph{Concept-Term List Expansion}
We begin with an expert-curated list of 77,000 legal and domain-specific concept-terms assembled by domain specialists. To increase coverage, we extract unique concept-terms from Named Entity Recognition datasets accumulated during prior information extraction work, expanding the list to 110,000 terms. This expanded list serves as the universe for all subsequent operations.

\paragraph{LLM-Augmented Dataset Generation}
We employ Gemini 2.5-Flash to generate semantic relationship labels through clustering-based prompt synthesis. The procedure operates as follows: (1) FastText Turkish embeddings (\texttt{cc\_tr\_300}) are computed for all 110,000 terms; (2) agglomerative clustering using cosine distance with threshold 0.4 produces 13,000 semantic clusters; (3) each cluster is submitted to Gemini 2.5-Flash with a system prompt requesting classification of intra-cluster relationships as Synonym, Antonym, or Co-hyponym. This synthetic generation yields approximately 827,000 labeled pairs at \$65 total cost.

\paragraph{Dictionary Integration}
To supplement synthetic generation with high-confidence examples, we incorporate a T\"{u}rk\c{c}e E\c{s} Anlaml{\i}lar S\"{o}zl\"{u}\u{g}\"{u} (Turkish Synonym Dictionary) containing approximately 20,000 words. Strict quality filtering retains only word entries having at most 2 synonyms per entry, contributing 16,000 high-confidence pairs. The combined dataset totals 843,000 labeled pairs with three-way categorical labels as shown in Table \ref{tab:dataset_specs}.

\subsection{Phase 2: Large-Scale Candidate Generation via GPU-Accelerated Vector Search}

Given 15 million terms and approximately 15 million existing embedding vectors, computing pairwise similarities naively would require $O(N^2)$ comparisons an intractable 225 trillion operations. We employ FAISS \cite{johnson2019billion} to reduce this to sublinear complexity through quantization and hierarchical indexing.

\paragraph{Vector Space Compression}
The 15 million embedding vectors (dimension 1024) require roughly 60 GB in full floating-point precision. To fit them within single-GPU VRAM limits, we apply 8-bit scalar quantization (SQ8), reducing the footprint to about 15 GB (a 4× compression) with minimal loss in accuracy:

\begin{equation}
q_i = \left\lfloor 127 \cdot \frac{v_i - \min_j v_j}{\max_j v_j - \min_j v_j} \right\rfloor
\end{equation}

\paragraph{Hierarchical Index Construction}
We construct an Inverted File (IVF) index partitioning the search space into 16,384 Voronoi cells via k-means clustering. At query time, only cells near the query vector are searched, reducing computation from $O(N)$ to approximately $O(\log N)$ cluster probes.

\paragraph{Similarity Search and Thresholding}
For each of the 15 million terms, we compute embeddings using the fine-tuned e5-large model and retrieve the top-k=100 neighbors by cosine similarity. We retain only pairs with cosine similarity exceeding 0.70, a permissive threshold intentionally chosen to prioritize recall over precision. This generates approximately 1.3 billion candidate pairs for Phase 3.

\subsection{Phase 3: Semantic Relationship Classification}

The 1.3 billion candidate pairs represent diverse semantic relationships beyond synonymy. To discriminate true synonyms from antonyms and co-hyponyms, we train multi-class classification models on the 843,000 labeled dataset.

\paragraph{Pair Classification Architecture}
We treat semantic relationship identification as a sentence pair classification task. Each input pair undergoes tokenization:

\begin{equation*}
\text{[CLS]} \oplus \text{term}_1 \oplus \text{[SEP]} \oplus \text{term}_2 \oplus \text{[SEP]}
\end{equation*}

The [CLS] token's final hidden representation is passed through a three-way linear classifier predicting classes: Synonym (label 2), Antonym (label 0), Co-hyponym (label 1). Maximum sequence length is set to 64 tokens.

\begin{figure*}[t]
\centering
\includegraphics[width=1.0\textwidth]{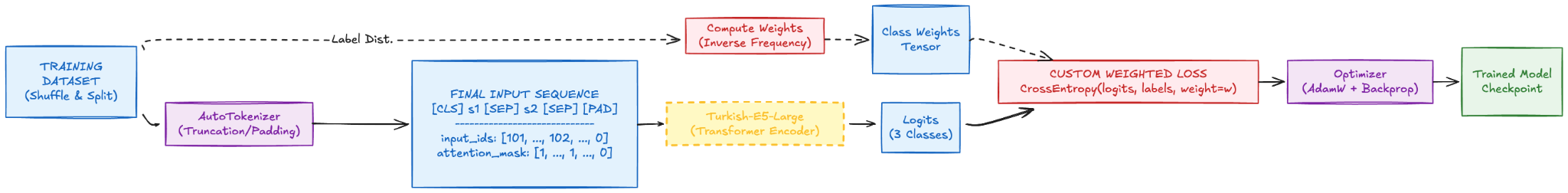}
\caption{Classification model training architecture and weighted loss calculation flow. Word pairs are tokenized and processed through a transformer encoder, with class weights computed from inverse frequencies to address label imbalance.}
\label{fig:classifier_arch}
\end{figure*}

\paragraph{Training with Weighted Loss}
The training architecture is illustrated in Figure~\ref{fig:classifier_arch}. We employ weighted CrossEntropy loss to address class imbalance:

\begin{equation}
\mathcal{L}_{\text{CE}} = -\sum_{c=1}^{3} w_c \cdot y_c \log(\hat{y}_c)
\end{equation}

where $w_c = N / (k \times n_c)$ assigns higher weights to underrepresented classes. Training uses batch size 128 with BF16 precision on NVIDIA L40S GPU.

\paragraph{Model Selection}
We evaluated six transformer-based sentence embedding models, namely TurkEmbed4STS~\cite{ezerceli2025turkembed}, modernbert-base-tr~\cite{newmindai2025modernbert}, mpnet~\cite{allmpnetbasev2}, bert-base-turkish-128k~\cite{bertbaseturkishcased}, turkish-e5-large~\cite{izdas2025turkish}, and multilingual-e5-large~\cite{wang2024multilingual} as shown in Table \ref{tab:candidates}. Based on superior F1-macro performance (0.87) and stable training dynamics, turkish-e5-large was selected for subsequent experiments.

\subsection{Phase 4: Filtering and Conflict Resolution}

\paragraph{Confidence and Class Filter}
We retain pairs satisfying: (1) classification label is ``Synonym''; (2) prediction confidence score exceeds 0.70. Pairs classified as Antonym or Co-hyponym are discarded.

\paragraph{Symmetry and Conflict Checking}
Semantic relationships should be approximately symmetric: if $(A, B)$ is a synonym pair, then $(B, A)$ should also be synonymous. We implement explicit symmetry validation by comparing predictions on both orderings. Pairs exhibiting conflict where one direction predicts Synonym but the reverse predicts Antonym are entirely removed. This filtering reduces the 1.3 billion candidates to approximately 520 million verified edges.

\subsection{Phase 5: Novel Clustering Algorithm for Semantic Drift and Polysemy Resolution}

The 520 million verified synonym pairs form a massive undirected graph with 15 million nodes. Standard community detection algorithms fail due to semantic drift and inability to handle polysemy.

\paragraph{Two-Stage Soft-to-Hard Architecture}
We replace standard algorithms with a two-stage hybrid approach: Stage 1 (Expansion) allows soft clustering via overlap-based membership; Stage 2 (Pruning/Reduction) resolves polysemy through topological voting.

\paragraph{Stage 1: Expansion and Soft Clustering}
Verified synonym pairs are sorted descending by confidence score, ensuring strongest bonds initialize clusters. These pairs are converted into an adjacency map. Cluster construction proceeds iteratively: a term $t$ joins cluster $C$ if the intersection ratio exceeds threshold:

\begin{equation}
\frac{|\text{synonyms}(t) \cap \text{members}(C)|}{|\text{members}(C)|} > 0.51
\end{equation}

This threshold prevents weak transitive chains while allowing legitimate multi-context membership. Critically, a term may satisfy this condition for multiple clusters, creating soft membership assignments.

\paragraph{Stage 2: Polysemy Resolution via Topological Voting}
Terms with multiple cluster memberships are resolved through a three-level hierarchical mechanism:

\begin{enumerate}
    \item \textbf{Majority Rule:} For each polysemous term, count shared synonyms with each candidate cluster. The cluster containing the most common synonyms wins: $\arg\max_C |\text{synonyms}(t) \cap \text{members}(C)|$.

    \item \textbf{Specificity Principle:} If clusters tie on majority rule, prefer the cluster with smaller cardinality: $\arg\min_C |C|$. This prevents absorption into massive ``catch-all'' clusters.

    \item \textbf{Determinism by ID:} If specificity also ties, assign to the cluster with smallest ID for reproducibility.
\end{enumerate}

\subsection{Phase 6: Parent Selection for Cluster Representation}

Each cluster requires a representative ``parent'' term serving as the canonical referent. We implement a two-stage selection pipeline:

\paragraph{Stage 1: Dictionary-Based Guarantee}
We maintain a curated list of 90,000 fundamental concept-terms. If any cluster member appears in this dictionary, that term is immediately designated parent without further computation.

\paragraph{Stage 2: Centroid-Based Selection}
For clusters lacking dictionary members ($\sim$97\% of clusters), we compute the semantic centroid by averaging L2-normalized embeddings of all members. The member with highest cosine similarity to this centroid becomes parent, representing the ``semantic center of gravity.''

\subsection{Phase 7: Final Output Generation}

The output obtained at the end of Phase 7 consists of a total of 2,905,071.454
semantic clusters. The median size of these
clusters is 3, and the mean cluster size is calculated as 4.58. The largest cluster size observed is 86. All of this structure has been generated in JSON format, which clearly represents the parent–child equivalence relationships.

\section{Experiments \& Results}

\subsection{Dataset}
\begin{table}[ht]
    \centering
    \caption{Dataset Specifications}
    \label{tab:dataset_specs}
    \small
        \begin{tabular}{|p{0.25\columnwidth}|p{0.65\columnwidth}|} 
        \hline
        \textbf{Component} & \textbf{Details} \\ 
        \hline
        \textbf{Total Size} & 843k unique pairs \\ 
        \hline
        \textbf{Data Sources} & Synthetic (Gemini 2.5-Flash): $\sim$827k \newline External Dictionary: $\sim$16k \\ 
        \hline
        \textbf{Labels} & Synonym, Antonym, Co-hyponym \\ 
        \hline
        \textbf{Format} & \texttt{\{"sentence1": "term\_A",} \newline \texttt{"sentence2": "term\_B", "label": "rel"\}} \\ 
        \hline
    \end{tabular}
\end{table}

\subsection{Experimental Setup}

\subsubsection{Phase 1: Model Candidate Selection}

Six transformer-based models were evaluated under identical conditions (Table~\ref{tab:candidates}).

\begin{table}[t]
\centering
\small
\caption{Candidate models for Phase 1 selection.}
\begin{tabular}{lcc}
\toprule
\textbf{Model} & \textbf{Arch.} & \textbf{Params} \\
\midrule
TurkEmbed4STS & GTE-multilingual-base & 305M \\
modernbert-base-tr & ModernBERT & 135M \\
mpnet & XLM-RoBERTa & 278M \\
bert-base-turkish-128k
 & BERT & 184M \\
turkish-e5-large  & XLM-RoBERTa & 560M \\
multilingual-e5-large & XLM-RoBERTa & 560M \\
\bottomrule
\end{tabular}
\label{tab:candidates}
\end{table}

All runs used identical hyperparameters on an NVIDIA RTX 3060, with a learning rate of $3 \times 10^{-5}$, a batch size of 64, and training conducted for 5 epochs using a maximum sequence length of 64. Computations were performed in BF16 precision, optimized with Fused AdamW, and a cosine learning rate scheduler with a warmup ratio of 0.1 was applied throughout training.

\subsubsection{Phase 2: Final Optimized Training}

Based on Phase 1 results, \texttt{turkish-e5-large} was selected for final training on upgraded hardware (NVIDIA L40S) with batch size increased to 128.

\subsubsection{Evaluation Strategy}

Model performance was monitored using F1-Macro score on the validation set. The \texttt{load\_best\_model\_at\_end=True} strategy ensured final checkpoints correspond to peak validation performance.

\subsection{Comparison of Experiments}

\begin{table}[t]
\centering
\caption{Classification model comparison. Phase 2 training on L40S with larger batch size yields best performance.}
\small
\begin{tabular}{lc}
\toprule
\textbf{Model} & \textbf{F1-Macro} \\
\midrule
TurkEmbed4STS & 0.82 \\
modernbert-base-tr & 0.79 \\
mpnet & 0.83 \\
bert-base-turkish-128k & 0.81 \\
turkish-e5-large (Phase 1) & 0.87\\
multilingual-e5-large & 0.84 \\
\midrule
\textbf{turkish-e5-large (Phase 2)} & \textbf{0.90}\\
\bottomrule
\end{tabular}
\label{tab:results}
\end{table}

\begin{table}[t]
\centering
\caption{Per-class performance of final model.}
\small
\begin{tabular}{lccc}
\toprule
\textbf{Class} & \textbf{Precision} & \textbf{Recall} & \textbf{F1} \\
\midrule
Synonym & 0.76& 0.90& 0.83\\
Antonym & 0.91& 0.93& 0.92\\
Co-hyponym & 0.93 & 0.95 & 0.94 \\
\midrule
\textbf{Macro Avg.} & 0.88& 0.92& \textbf{0.90}\\
\bottomrule
\end{tabular}
\label{tab:perclass}
\end{table}

\begin{figure*}[t]
    \centering
    \begin{subfigure}[b]{0.48\textwidth}
        \includegraphics[width=\linewidth]{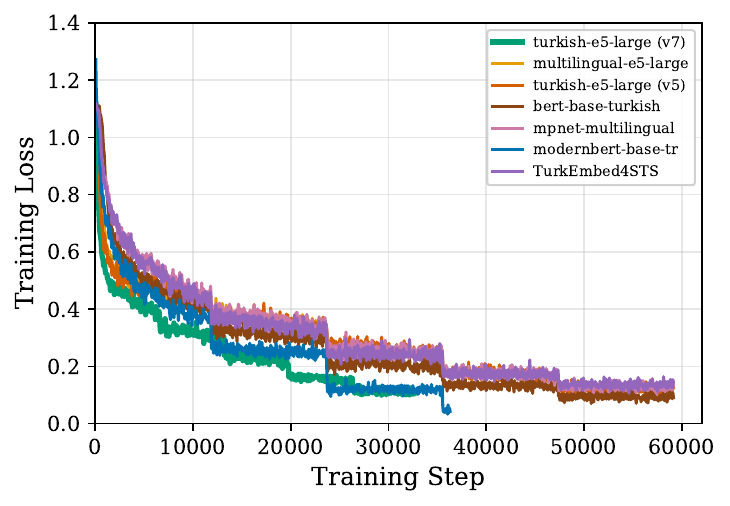}
        \caption{Training Loss}
    \end{subfigure}
    \hfill
    \begin{subfigure}[b]{0.48\textwidth}
        \includegraphics[width=\linewidth]{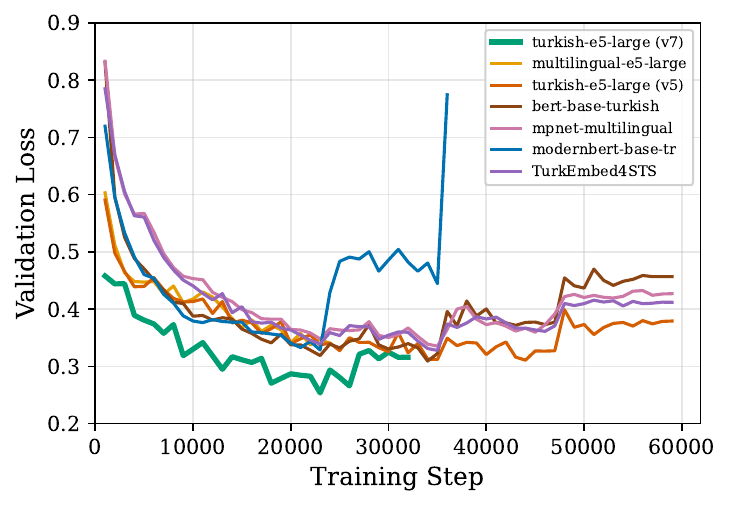}
        \caption{Validation Loss}
    \end{subfigure}
    
    \vspace{0.3cm}
    
    \begin{subfigure}[b]{0.48\textwidth}
        \centering
        \includegraphics[width=\linewidth]{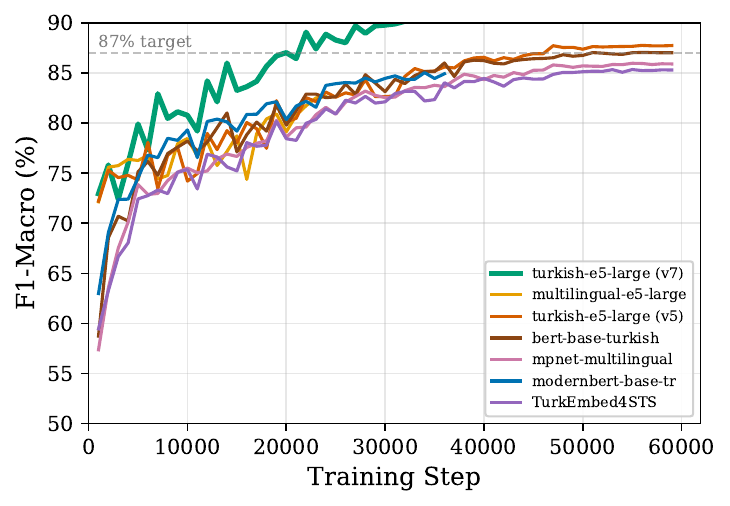}
        \caption{F1-Macro Score}
    \end{subfigure}

    \caption{Training dynamics of the semantic relation classifier. Subfigures (a) and (b) demonstrate the convergence of cross-entropy loss on training and validation sets, respectively. Subfigure (c) shows the corresponding rise in F1-Macro scores, where the \texttt{turkish-e5-large} variant (green line) demonstrates superior efficiency, reaching a peak score of 0.90 significantly faster than competing baselines.}
    \label{fig:training_wide}
\end{figure*}

The turkish-e5-large model demonstrated superior performance and stability across training runs (Figure~\ref{fig:training_wide}). The final model achieves an overall \textbf{90\% F1-Macro} score (in Table \ref{tab:results}), with \textbf{94\% F1} on co-hyponym classification, \textbf{92\% F1} on antonym detection, and \textbf{83\% F1} on synonym identification as in Table \ref{tab:perclass} and Appendix~\ref{app:metrics}.

\subsection{Clustering Results}

Our synonym graph construction pipeline processed \textbf{15 million} unique terms, generating approximately \textbf{1.3 billion} candidate term pairs for verification. After semantic filtering, \textbf{520 million} pairs were confirmed as valid synonym relationships, resulting in \textbf{2,905,071} final semantic clusters. The resulting clusters are compact and well-balanced, with a \textbf{median size of 3}, a \textbf{mean size of 4.58}, and a \textbf{maximum cluster size of 86}, indicating limited semantic drift while maintaining broad lexical coverage.

\subsection{Qualitative Analysis: Parent Selection}

Our parent selection demonstrates robust handling of complex linguistic phenomena:

\paragraph{OCR Error Correction}
Cluster [``M\"{u}cbir Sebe'', ``M\"{u}cbir Sebep Halleri'', ``Mucbir Sebepler'', ``M\"{u}cbir Sebep''] correctly selects parent ``M\"{u}cbir Sebep'' (Force Majeure), eliminating truncated OCR output ``Sebe'' and diacritic errors ``Mucbir''.

\paragraph{Formal Terminology Priority}
Cluster [``VUK'', ``Vergi Usul K.'', ``213 Say{\i}l{\i} Kanun'', ``Vergi Usul Kanunu'', ``Vergi Usul Yasas{\i}''] correctly identifies ``Vergi Usul Kanunu'' (Tax Procedure Code) as parent rather than abbreviation ``VUK''.

\paragraph{Institutional Name Standardization}
Cluster [``SSK'', ``Sosyal Sigortalar'', ``SGK'', ``Sosyal G\"{u}venlik Kurumu'', ``Sosyal G\"{u}venlik Te\c{s}kilat{\i}''] correctly selects ``Sosyal G\"{u}venlik Kurumu'' (Social Security Institution, current official name).

\section{Discussion and Limitations}

\subsection{Key Findings}

Our experiments demonstrate several important findings:

\paragraph{LLM-Generated Labels Are Reliable}
The 90\% F1-macro achieved by classifiers trained on Gemini 2.5-Flash labels indicates that LLM-generated semantic relationships exhibit sufficient quality for downstream applications. This validates the cost-effective approach of synthetic data generation over manual annotation.

\paragraph{Turkish-Specialized Models Outperform Multilingual Baselines}
The \texttt{turkish-e5-large} model, pre-trained on Turkish text, outperforms the multilingual model multilingual-e5-large despite identical architecture, suggesting that language-specific pre-training remains valuable even for well-resourced multilingual models.

\paragraph{Soft-to-Hard Clustering Resolves Polysemy}
Our two-stage clustering algorithm successfully handles polysemous terms that would be misclassified by traditional hard-partitioning algorithms. The topological voting mechanism provides principled disambiguation while maintaining computational tractability.

\paragraph{Class Distribution and Performance Analysis}
The dataset exhibits a notable imbalance, with the co-hyponym class being dominant in terms of sample count. This class distribution directly accounts for the 94.8\% F1-score on co-hyponyms, as the model learns this relation from abundant training examples. Interpreting the synonym class’s 83.1\% F1-score, however, requires situating it within the central motivation of this study. The primary objective is not simply to distinguish synonyms from antonyms, but to separate synonyms from co-hyponyms. This distinction matters because co-hyponyms share hypernyms and often exhibit high semantic similarity, yet they encode a fundamentally different relation. As a result, their embeddings tend to cluster near synonym pairs, which makes discrimination substantially more challenging than separating synonyms from antonyms that occupy clearly opposing regions of semantic space.

The model's ability to achieve 90\% macro F1-score while successfully separating these semantically proximate classes demonstrates that it has learned to model not just semantic similarity, but also the conceptual specificity required to distinguish ``same meaning'' (synonymy) from ``same category'' (co-hyponymy). This is precisely the capability needed for constructing antonym-free synonym graphs that avoid false merging of related but non-synonymous terms.







\section{Conclusion}

We present a comprehensive system for constructing antonym-free synonym graphs at massive scale, addressing the fundamental challenge that distributional similarity conflates synonymy with other semantic relationships. Our three contributions a large-scale disambiguated dataset, a specialized discriminator achieving 90\% F1-macro, and a novel soft-to-hard clustering algorithm together enable the construction of the largest native Turkish semantic resource to date: 2.9 million synonym clusters derived from 15 million terms.

The resulting resource enables high-precision semantic search and retrieval-augmented generation for morphologically rich languages where existing synonym databases remain sparse. Our methodology is designed for cross-linguistic transfer, requiring only FastText embeddings, LLM API access, and basic dictionary resources components available for hundreds of languages.

Future work includes extending the approach to additional morphologically rich languages, incorporating explicit morphological paradigm expansion, and developing dynamic update mechanisms to maintain temporal validity as terminology evolves.








\bibliography{custom}

\appendix
\section*{Appendices}
\section{Additional Examples}
\label{app:examples}

\paragraph{Semantic Drift Prevention}
Without our filtering, the chain ``S{\i}cak'' (Hot) $\rightarrow$ ``Ac{\i}'' (Spicy) $\rightarrow$ ``Ac{\i}'' (Pain) $\rightarrow$ ``\"{U}z\"{u}nt\"{u}'' (Sadness) $\rightarrow$ ``Depresyon'' (Depression) would create a single cluster conflating temperature with mental health. Our confidence-based filtering and intersection ratio thresholds prevent such spurious chaining.

\paragraph{Polysemy Handling}
The term ``y\"{u}z'' (face/100) is initially soft-assigned to both an anatomical cluster (containing ``\c{c}ehre,'' ``surat,'' ``sima'') and a numerical cluster (containing ``y\"{u}zer,'' ``y\"{u}zde''). Topological voting assigns it to the anatomical cluster based on higher synonym overlap, while the numerical sense is captured through related terms.

\section{Detailed Training Metrics}
\label{app:metrics}

\begin{figure}[hp]
    \centering
    \begin{subfigure}[b]{\columnwidth}
        \centering
        \includegraphics[width=\linewidth]{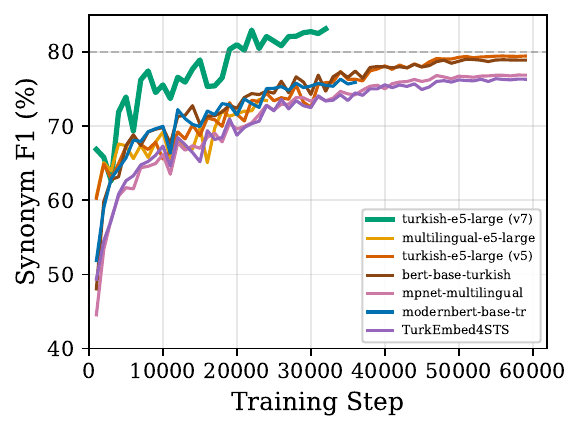}
        \caption{Synonym F1}
    \end{subfigure}
    
    \vspace{0.3cm} 
    
    \begin{subfigure}[b]{\columnwidth}
        \centering
        \includegraphics[width=\linewidth]{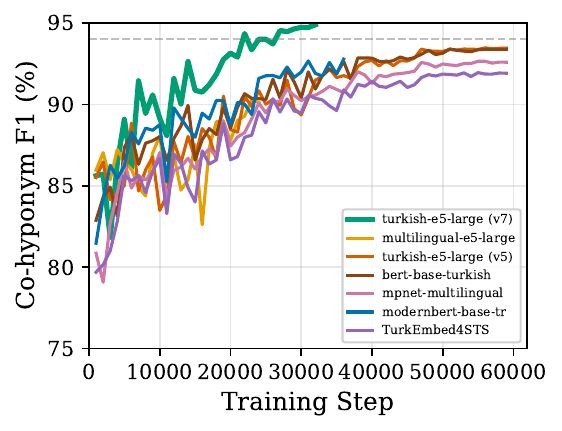}
        \caption{Co-hyponym F1}
    \end{subfigure}
    
    \caption{Per-class F1 scores during training. Co-hyponym classification converges quickly to 0.94, while synonym detection shows more variation across models, stabilizing at 0.83 for the best model.}
    \label{fig:perclass_f1}
\end{figure}

\section{Algorithm Pseudocode}
\label{app:algorithm}

\begin{algorithm}[H]
\caption{Soft-to-Hard Clustering}
\begin{algorithmic}[1]
\Require Synonym pairs $P$ with confidence scores
\Ensure Final hard clusters $C$
\State Sort $P$ by confidence descending
\State Initialize cluster set $C \gets \emptyset$
\State Initialize membership map $M$
\For{each pair $(u, v, \text{conf})$ in $P$}
    \State Find candidate clusters for $u$ and $v$
    \If{both unassigned}
        \State Create new cluster $c = \{u, v\}$
        \State $C \gets C \cup \{c\}$
    \ElsIf{intersection ratio $> 0.51$}
        \State Add unassigned term to existing cluster
        \State Update membership map $M$
    \EndIf
\EndFor
\For{each term $t$ with $|M[t]| > 1$}
    \State Apply topological voting:
    \State \quad 1. Majority rule
    \State \quad 2. Specificity principle
    \State \quad 3. Determinism tiebreak
    \State Assign $t$ to winning cluster
\EndFor
\State \Return $C$
\end{algorithmic}
\end{algorithm}

\end{document}